%% file: main.tex
\documentclass[10pt, a4paper]{article}

\usepackage[final]{lrec-coling2024} % this is the new style

\usepackage{inconsolata}
\usepackage{amsmath}
\usepackage{graphicx}
\usepackage{hyperref}
\usepackage[toc,page]{appendix}
\usepackage{svg}
\usepackage{amssymb}
\usepackage{booktabs} 
\usepackage{tcolorbox}
\usepackage{xcolor}

\title{Robust and Scalable Model Editing for Large Language Models}
\name{
    Yingfa Chen$^{1}$,
    Zhengyan Zhang$^{1}$,
    Xu Han$^{1,3,\dagger}$,
    Chaojun Xiao$^1$,
    Zhiyuan Liu$^{1,\dagger}$,\\
    \fontsize{12}{14}\selectfont\textbf{Chen Chen$^{2}$,
    Kuai Li$^{2}$,
    Tao Yang$^{2}$,
    Maosong Sun$^{1}$}
}

\address{
    $^{1}$DCST, IAI, BNRIST, Tsinghua University, Beijing, China \\
    $^{2}$Tencent Machine Learning Platform, China\\
    $^{3}$Shanghai Artificial Intelligence Laboratory\\
    yf-chen22@mails.tsinghua.edu.cn\\
    \{hanxu2022, liuzy\}@tsinghua.edu.cn
}

\abstract{
Large language models (LLMs) can make predictions using \textit{parametric knowledge}--knowledge encoded in the model weights--or \textit{contextual knowledge}--knowledge presented in the context. In many scenarios, a desirable behavior is that LLMs give precedence to contextual knowledge when it conflicts with the parametric knowledge, and fall back to using their parametric knowledge when the context is irrelevant. This enables updating and correcting the model's knowledge by in-context editing instead of retraining. Previous works have shown that LLMs are inclined to ignore contextual knowledge and fail to reliably fall back to parametric knowledge when presented with irrelevant context. In this work, we discover that, with proper prompting methods, instruction-finetuned LLMs can be highly controllable by contextual knowledge and robust to irrelevant context. Utilizing this feature, we propose EREN (Edit models by REading Notes) to improve the scalability and robustness of LLM editing. To better evaluate the robustness of model editors, we collect a new dataset, that contains irrelevant questions that are more challenging than the ones in existing datasets. Empirical results show that our method outperforms current state-of-the-art methods by a large margin. Unlike existing techniques, it can integrate knowledge from multiple edits, and correctly respond to syntactically similar but semantically unrelated inputs (and vice versa).
The source code can be found at \url{https://github.com/thunlp/EREN}.
 \\ \newline  }

\begin{document}

\maketitleabstract

{\let\thefootnote\relax\footnotetext{$^\dagger$ Corresponding authors}}

\input{contents/intro}
\input{contents/related_works}
% \input{contents/problem}
\input{contents/method}

\input{contents/experiments}
\input{contents/result}

\input{contents/conclusion}

\section*{Acknowledgement}

This work is supported by the National Key R\&D Program of China (No. 2022ZD0160501) and National Natural Science Foundation of China (No. 62236011, No. 62236004). 
This work is also supported by Tencent AI Platform Department.

The method is primarily designed by Yingfa Chen, and he conducted most of the experiments. Zhengyan Zhang, Xu Han, and Chaojun Xiao, Zhiyuan Liu actively participated in experimental design, result analysis, and paper writing. Chen Chen, Kuai Li, Tao Yang, and Maosong Sun provided valuable advice to this research.

\nocite{*}
\section{Bibliographical References}\label{sec:reference}

\bibliographystyle{lrec-coling2024-natbib}
\bibliography{custom}
% \bibliography{lrec-coling2024-example}

% \section{Language Resource References}
% \label{lr:ref}
% \bibliographystylelanguageresource{lrec-coling2024-natbib}
% \bibliographylanguageresource{languageresource}

\input{contents/appendix}

\end{document}

%% file: contents/intro.tex
\section{Introduction}

Large language models (LLMs) have demonstrated remarkable performance on numerous natural language processing (NLP) tasks and can memorize vast amounts of knowledge \cite{lm-as-kb,autoprompt,gpt3,cbqa,gpt4}. 
However, the memorized knowledge may not be consistent with the knowledge of the real world \cite{temporal-generalization,temporal-kb}, and this may lead to undesired behaviors or incorrect predictions \cite{hallucination-survey}.
Hence, model editing~\citep{enn,zhu2021modifying}, which aims to quickly modify the behavior of a deployed LLM on specific examples while preserving its performance on unrelated instances, has gained attention in recent years.

The early approaches to model editing have focused on direct updates to the model parameters~\citep{enn,zhu2021modifying,decao-editing,mend,knowledge-neurons,rome,memit,controllable-working-mem}, which cannot be applied to current LLMs due to the inaccessibility of the parameters~\citep{gpt4,palmv2}. 
Recently, some preliminary studies have explored the possibility of in-context model editing~\citep{prompting-gpt3}, which modifies the behavior of a deployed LLM by adding a prompt to the input. 
The concurrent work by \citet{ike} builds upon this by adding demonstrations for behavior preservation on unrelated edit examples.
% zzy: 引用一些相关的工作，比如北大那个之类的，下面一段开头的地方类似
In this way, the model users can easily edit black-box LLMs without access to the model parameters.

\citet{controllable-working-mem} shed light on a more scalable in-context model editing method. They argue that we can view the context as the \textit{working memory}~\citep{working-mem,Ashby_Ell_Valentin_Casale_2005} of neural models, and propose a finetuning regime that drives an LLM to make predictions grounded on the knowledge presented in the context over the knowledge it has learned during pretraining. 
% However, it has been shown that LLMs have poor controllability and robustness, namely, the LLM sometimes ignore the knowledge presented in the context, or fails to ignore irrelevant knowledge, negatively impacting their result~\citep{controllable-working-mem,kalm,llm-irrelevant-context}. Moreover, their method requires access to the LLM's parameters. 

However, existing in-context model editing methods~\citep{prompting-gpt3,ike,controllable-working-mem} have three major limitations.
(1)~They are not scalable to large numbers of edits. If we integrate multiple edits into a single prompt, the prompt may be too long for the LLM. 
% Besides, the LLM may preferentially focus on the last edits in the prompt, and ignore the previous edits~\citep{DBLP:journals/csur/LiuYFJHN23}. 
% zzy: 这里的引用可以调研一下，我没有仔细找，最好是有一些论文直接和这些论点有关系，可以支撑
(2)~They assume the relevant edit of a certain instance is given while in real-world scenarios the model needs to determine whether the current instance is related to any edits.
If an instance is unrelated to all edits but we still use the edits as prompts, it often has a negative impact~\citep{DBLP:conf/emnlp/JiaL17,DBLP:journals/corr/abs-2301-07085}.
(3)~LLMs sometimes ignore the knowledge presented in the context or fail to ignore irrelevant knowledge, negatively impacting their result~\citep{controllable-working-mem,kalm,llm-irrelevant-context}. 

% \citet{controllable-working-mem} argues that we can view the neuron activations as the working memory~\citep{working-mem,Ashby_Ell_Valentin_Casale_2005} of neural models, and propose a finetuning regime that drives an LLM to make predictions using knowledge in the context over the knowledge it has learned during pretraining. However, it has been shown that LLMs have poor controllability and robustness, namely, the LLM sometimes ignore the knowledge presented in the context, or fails to ignore irrelevant knowledge, negatively impacting their result~\citep{controllable-working-mem,kalm,llm-irrelevant-context}. Moreover, their method requires access to the LLM's parameters. 

In this work, we show that instruction-tuned LLMs can be reliably grounded on contextual knowledge. Inspired by this, we propose a robust and scalable model editing method called EREN (Edit models by REading Notes).
(1)~Specifically, the LLM is complemented with a notebook memory that stores all edits in natural text. 
For a given input, relevant edits are retrieved from the notebook and used as prompts to modify the behavior of the LLM.
In this way, we can easily scale up the number of edits without increasing the length of the prompt.
(2)~To determine whether the current instance is related to a certain edit, we reformat the task of model editing into reading comprehension with an ``unanswerable'' option.
% For the instances that are not related to any edit, the model will predict ``unanswerable'' and re-generate the answer without any edit.
Hence, we can avoid the negative impact of irrelevant edits on the LLM behavior.

% In the experiments, we evaluate our method on instruction-tuned LLMs~\citep{instructgpt,flan,flan2}, which can be applied to a wide range of tasks by simply adding a few instructions to the input and have the ability to conduct reading comprehension based on edit prompts.
Empirical results show that our method can achieve state-of-the-art performance on in-context model editing on question answering and fact-checking.

Our main contributions are as follows:
\begin{itemize}
    \item We conduct rigorous experiments to show that instruction-tuning enables LLMs to give precedence to contextual knowledge over parametric knowledge.
    \item We propose EREN, a robust and scalable in-context model editing method that can handle large numbers of edits and irrelevant edits. Our method beats the current state-of-the-art by a large margin. % zzy: 这里好像不太好说multiple edits，类似abstract里提到的问题
    % We propose a model editing method that does not require any training, parameter update, or labeled data, and it beats the current state of the art by a large margin. It is also the only retrieval-based method that can condition answers on multiple edits. 
    % \item We evaluate our method on model sizes from less than 100M to more than 10B, and show that the performance of our method increases predictably with model size.
    \item We process and release cleaner and more challenging versions of existing datasets for model editing, and empirically show that existing methods see drastic performance drops on our new types of challenging examples.
\end{itemize}

%% file: contents/related_works.tex
\section{Related Works}

\paragraph{Model Editing}

Our method is most related to the lines of work on the model editing problem. It was originally proposed by \citet{enn,zhu2021modifying}. \citet{enn} proposes a meta-learning framework to train models that can be more easily edited. \citet{zhu2021modifying} uses $L_p$ norm to constrain the parameter change while training the model. KnowledgeEditor \citep{decao-editing}, MEND \citep{mend}, and \citet{cross-lingual-model-editing} introduce hyper-networks to transform gradients into parameter changes. 
% \citet{cross-lingual-model-editing} addresses model editing in a multi-lingual setting. 
However, the performance of gradient-based methods suffers greatly when applying multiple edits in sequence, and gradient information may be unavailable.

ROME \citep{rome} and MEMIT \citep{memit} propose causal tracing to locate factual associations in a GPT \citep{gpt2,gpt-j}, and update the FFN layer to insert a factual association. However, these method requires expensive activation statistics and do not work with non-causal LMs. REMEDI \citep{remedi} proposes to use learn a mapping from inputs to the hidden representations to guide the output, but it only focuses on editing errors in the input. 

Retrieval methods use an external memory module and an explicit relevance estimation step to avoid reliance on gradient information or knowledge-locating methods. 
SERAC \citep{serac} estimates relevance using a scope classifier and sends relevant edits to a counterfactual model. 
GRACE \citep{grace} caches and retrieves hidden representations of edits. 
However, these methods have limitations in generalization and performance, and assume that there is only one relevant edit at a time.
In contrast, EREN's relevance estimation is more accurate, can be conditioned on multiple edits at the same time, and has stronger generalization abilities.

Finally, the concurrent work by \citet{ike} uses in-context learning to update the model's knowledge. Still, they only consider the insertion of one fact and assume that the relevant fact is known. Their method can be seen as the few-shot version of our one-step MRC baseline. 

\paragraph{Retrieval-Augmented Methods}

Our work relies on retrieval to scale up to thousands of edits. Retrieval-augmented methods have demonstrated impressive capabilities in knowledge intensive tasks~\citep{chen-reading-wikipedia,dpr,rag,realm,replug}. More recent and concurrent works include \citep{llm-knowledge-boundary,freshllm,Yoran_Wolfson_Ram_Berant_2023,kalm,Zhou2023ContextfaithfulPF,flare,llm-irrelevant-context}.

\paragraph{Prompting with Retrievals}

Our methodology can be categorized as ``prompting''. MemPrompt \citep{memprompt} employs a growing memory of prompts to help the model better understand user intentions, but they focus on a different task setting. \citet{prompting-gpt3} prompts GPT-3 to perform reading comprehension on Wikipedia passages with replaced entities to update knowledge, but have little analysis related to model editing. 
% Additionally, updating lengthy Wikipedia archives can also be challenging in practice. Finally, \citet{generative-agents} have also analyzed using a dynamic memory to store text-based experiences for autonomous agents. Such information can be recorded in the notebook of EREN.

%% file: contents/method.tex
\section{Methodology}

% \subsection{Problem Formulation}

\begin{figure*}[ht!]
    \centering
    \includegraphics[width=0.9\textwidth]{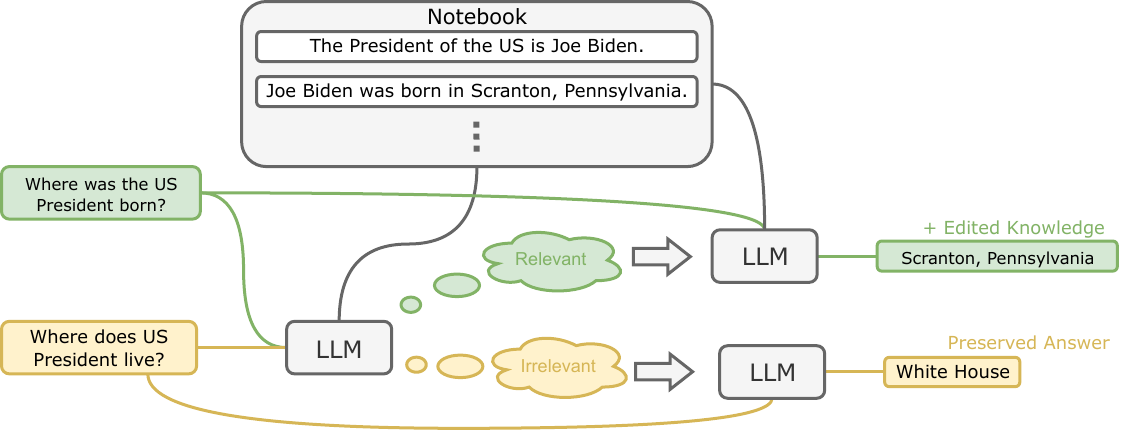}
    \caption{Illustration of the framework of EREN. Two edits have been injected, and the colored part shows inference on two inputs. Green: Both edits are relevant, and the final output depends on both. Yellow: The LLM determines that no edit is relevant, and the output of the base model is used.}
    \label{fig:framework}
\end{figure*}

\subsection{Sequential Model Editing}
\label{sec:model-editing-problem}

In model editing, multiple edits may be applied simultaneously or sequentially. The ability to perform the latter is important for making the edits as soon as they appear and has been shown to be considerably more challenging than the former scenario~\citep{t-patcher}. This paper focuses on sequential model editing.

When we apply one \textit{edit} $e$, we want to instill certain behaviors into a model $f$ on a set of inputs. Typically, $e$ is a fact and we want the edited model $f^*$ to behave as if the fact is true. The goal of model editing is to find an \textit{editor} function that produces an edited model given a \textit{base model} and an edit: $\text{Edit}(f, e) = f^*$.

To evaluate the correctness of $f^*$, we define the \textit{edit scope} $I(e)$ of $e$ as the set of input-output pairs that are implied by $e$:
\begin{equation*}
    I(e)=\{(x_1, y_1), ..., (x_m, y_m)\},
\end{equation*}
where $(x_i, y_i)$ is the $i$-th example implied by $e$.

For instance, the edit that instills the fact that ``The CEO of Apple is Tim Cook'' implies that the answer to the questions ``Who is the CEO of Apple?'' and ``Where does Tim Cook work?'' are ``Tim Cook'' and ``Apple'', respectively. 

In \textit{sequential} model editing, we want to apply each edit one by one and ensure that all intermediate edited models are performant. This is important for keeping the model up-to-date and undesired behaviors are fixed as soon as possible. Assume an ordered set of $n$ edits $\mathcal{E} = \{ e_1, ..., e_n \}$, the edited model is as follows.\footnote{$f_1 \circ f_2$ denotes the composite of $f_1$ and $f_2$.}
\begin{equation*}
\begin{small}
    f^* = \text{Edit}(f, \mathcal E) = \text{Edit}_n \circ \dots \circ \text{Edit}_2 \circ \text{Edit}_1 (f),
\end{small}
\end{equation*}
where $\text{Edit}_i(\cdot)=\text{Edit}(\cdot, e_i)$. Moreover, the edit scope of multiple edits is not the union of their scopes, because multiple edits in conjunction may imply new input-output pairs. Figure \ref{fig:framework} shows one such example in green. To address this, we denote the edit scope of a set of edits $I(\mathcal E)$ as all input-output pairs implied by all the edits in $\mathcal E$ in conjunction.%, and similarly for $O(\mathcal E)$.

The goal of $\text{Edit}(f,\mathcal {E})$ is to produce an $f^*$ that satisfies the following.
\begin{align*}
    f^*(x) = \begin{cases}
        y    & \forall(x, y) \in I(\mathcal E) \\
        f(x) & \forall x     \not \in I(\mathcal E).
    \end{cases}
\end{align*}
% where $\overline{\cdot}$ denotes the complement of a set.

For the simplicity of further discussion, we say an edit is \textit{relevant} to an input $x$ (and vice versa) when its edit scope contains $x$.

Note that the edits may come in different formats. Most existing works on model editing, represent edits with input-output pairs \cite{enn,mend,serac}, while others use factual triples \cite{rome,memit,remedi}. However, this paper assumes that edits are given as short declarative sentences. 
% We believe this is a more natural manner to "tell" the model the desired behavior because declarative sentences are the most common type of sentence.

\subsection{Our Approach}

In summary, the edited LLM is complemented with a notebook that caches all edits in natural text. For each question, the model first determines whether the input is relevant to any edit, if so, it makes a prediction based on the notebook. Else, it directly answers the question using its memorized knowledge.

We find that LLMs, even when instruction-finetuned, are not readily controllable by their context, i.e. the notebook. In particular, they are not robust to irrelevant context~\citep{controllable-working-mem,kalm,llm-irrelevant-context}, resulting in changed predictions on unrelated inputs. Also, the number of edits may to too large to fit into the input context of the LLM. Addressing these two issues, we propose to (1)~split inference into two steps, and (2)~use a dual-encoder retrieval framework to perform rough relevance estimation.

\subsubsection{Two-Step Inference}

Our preliminary experiments reveal that LLMs, regardless of whether it is instruction-finetuned, are generally easily controllable by grounding on relevant contexts, but they are not robust to irrelevant context, which has been highlighted by existing works~\citep{controllable-working-mem,llm-irrelevant-context}. However, we discover that \textbf{instruction-finetuned LLMs can reliably determine the relevance of contexts.}

Inspired by this observation, we design a two-step inference pipeline. The LLM is first prompted to determine whether an input is relevant to existing edits, i.e., determine whether $x \in I(\mathcal E)$ is true. If true, the LLM performs conditional generation with all edits as the premise. If false, the LLM answers without context. One possible prompt template for relevance estimation is roughly as follows. 
The complete prompts are given in Appendix~\ref{appendix:prompts}.
% The complete prompts will be given in the appendix.
\begin{tcolorbox}
    \texttt{Read this and answer the question. If it is unanswerable, say \textcolor{red}{<irr>}.
    \\
    \textcolor{orange}{<premise>}
    \\
    \textcolor{blue}{<question>}
    }
\end{tcolorbox}

Here, \textcolor{orange}{\texttt{<premise>}} is a list of edits, and \textcolor{red}{\texttt{<irr>}} is a special token that indicates irrelevance. If the LLM's answer is \textcolor{red}{\texttt{<irr>}}, we prompt it to answer using only parametric knowledge.

In summary, the edited LLM can be formalized as:
\begin{align*}
    f^*(x) = \begin{cases}
        f(x)                                    & f(\mathcal T_\text{rel}(x, \mathcal E)) = \text{\textcolor{red}{\texttt{<irr>}}} \\
        f(\mathcal T_\text{gen}(x, \mathcal E)) & \text{otherwise},
    \end{cases}
\end{align*}
where $\mathcal T _\text{gen}$ and $\mathcal T _ \text{rel}$ are the prompt templates for conditional generation and relevance estimation.

This is analogous to a person noting down every edit, and using relevant notes over memorized knowledge if there are any relevant notes. 
Thus, we named the method EREN (Edit models by REading Notes). The framework is illustrated in Figure~\ref{fig:framework}.

\subsubsection{Rough Relevance Estimation}

In practice, $\mathcal T_\text{gen}(x, \mathcal E)$ and $\mathcal T_\text{rel}(x, \mathcal E)$ becomes exceedingly long when $\mathcal E$ is very large, and the input length may exceed the context capacity. Therefore, we perform a rough relevance estimation to eliminate irrelevant edits that are easily identified. To this end, an embedding-based \textit{note retriever} is employed to retrieve the top-$k$ most relevant notes. Let $R$ denote the encoder, the retrievals are
\begin{align*}
    \mathcal E_R = \text{Top-}k_{e \sim \mathcal E} \left( R(x) \cdot R(e) \right)
\end{align*}
where $k < |\mathcal E|$. We use $\mathcal E_R$ instead of $\mathcal E$ to construct the prompt.

%% file: contents/experiments.tex
\begin{table*}[!ht]
    \small
    \centering
    \begin{tabular}{p{20mm}|p{64mm}|p{60mm}}
        % \hline
        \toprule
        \textbf{Part} 
            & \textbf{Explanation} & \textbf{Example} \\
        % \hline
        \midrule
        Edit statement
            & A statement of the fact to be inserted.
            & ``The president of the US is Joe Biden.'' \\
        % \hline
        \midrule
        Edit scope    
            & QA pairs that are implied by the edit. 
            & ``Who is the president of the US?'', ``Joe Biden'' \\
        % \hline
        \midrule
        Out-of-scope examples
            & Questions whose answers are not changed by the edit.
            & ``Where does the president of the US live in?'', ``White House'' \\
        % \hline
        \bottomrule
    \end{tabular}
    \caption{Parts of an example in our version of \textsc{CounterFact}. The edit statement is only used for our method, and we use one QA pair from the edit scope for baseline methods that rely on QA pairs as edits.}
    \label{tab:cf-example}
\end{table*}

\section{Experiments}
\label{sec:experiments}

\subsection{Datasets}
\label{sec:datasets}

We evaluate the editors on QA and fact-checking. 
We first collect more challenging examples, then filter out examples of poor quality, and finally perform the necessary conversions to suit our setting.

\subsubsection{Question Answering}
\label{sec:dataset-qa}

% QA aims to answer questions where no context is provided. 
For QA, We use \textsc{CounterFact} \citep{rome}, a dataset for editing knowledge in language models. Each question in \textsc{CounterFact} is the verbalization of a factual triple (subject, relation, object), and edits are created by modifying the object. \textsc{CounterFact} also includes out-of-scope inputs (i.e., inputs outside of the edit scope) that are constructed by swapping the subject with a neighboring subject (see \citet{rome} for more details). 

While many existing model editors are evaluated on ZsRE \citep{zsre}, we choose \textsc{CounterFact} over it because the out-of-scope examples in ZsRE are sampled from a large set of unrelated examples, which are syntactically very different from the edit, making the edit scope estimation overly simple~\citep{rome}. Table~\ref{tab:cf-example} shows an example in our version of \textsc{CounterFact}.

% Moreover, we wish to compare our method against SERAC \citep{serac}, a state-of-the-art retrieval-based method, but it is trained with ZsRE, so we cannot use it because we are interested in the zero-shot setting.

\paragraph{Collecting Harder Out-of-Scope Questions}

We find that out-of-scope examples constructed by keeping the subject but changing the relation and object are more challenging. We hypothesize that this is because existing methods are overly reliant on the subject. Therefore, to better evaluate \textit{specificity} of model editors, i.e., their performance on out-of-scope questions, we generate such examples by collecting Wikidata triples with the same subject but different relations and objects, then verbalize them with the templates from \textsc{CounterFact}.

% \paragraph{Conversion}

% Inputs in \textsc{CounterFact} are designed for text completion, but we primarily test encoder-decoder models, so we convert prefix prompts to QA pairs using MixQG~\citep{mixqg}. Each example $(e, I(e), O(e))$ includes the edit, the edit scope, and out-of-scope examples. Table \ref{tab:cf-example} shows the different parts of an example in our version of \textsc{CounterFact}.
See Appendix \ref{appendix:dataset-construction} for more details.

% Note that the edit scope in \textsc{CounterFact} is created by simply paraphrasing one question, which means all answers are the same. 
% An ideal edit scope for evaluation should include all QA pairs implied by the edit, but such data is too expensive to collect. 
% Thus, we leave such study for future work.

\subsubsection{Fact-Checking}
\label{sec:fever}

We follow existing works \citep{decao-editing,serac} and evaluate using FEVER~\citep{fever} where each example is a factual statement. We use the version released by \citet{decao-editing} which includes input paraphrases generated with back-translation. 
The fact statement itself is used as the edit statement. %, which means that the edited model should regard the fact statement as true. 
The reader performs natural language inference (NLI) with the retrieved edits as premises. For editors that require QA pairs, we convert the facts into boolean questions with the template ``Is it true that \{statement\}?''. To generate false facts, we sample half of all facts, and flip the answers to all questions except the first one to ``no'' by negating the fact statements using a BART-Base \citep{bart} from \citet{crossaug}.

\subsubsection{Filtering} 
\label{sec:filtering}

\begin{table}[t!]
    \small
    \centering
    \begin{tabular}{lcc}
        % \hline
        \toprule
        \textbf{Version} & \textsc{CounterFact}  & FEVER \\
        % \hline
        \midrule
        Original       & $12.5 \%$ / $36.9\%$ & $42.7 \%$ \\
        Auto-filtered  & $0 \%$    / $ 0\% $  & $26.5 \%$ \\
        % \hline
        \bottomrule
    \end{tabular}
    \caption{Proportion of incorrectly labeled examples in \textsc{CounterFact} and FEVER, by human inspection on 128 samples. For \textsc{CounterFact}, the two numbers correspond to the error proportion of in-scope and out-of-scope examples.}
    \label{tab:bad-example-breakdown}
\end{table}

The datasets have a relatively large proportion of erroneous labels, and we leverage pretrained models to create filtered versions.
Table \ref{tab:bad-example-breakdown} shows the proportion of erroneous labels in the original and filtered versions. However, we find that the auto-filtered FEVER still contains many erroneous examples, so we create a cleaner version of FEVER by manual filtering with 128 examples. More details about the implications of filtering and some examples of erroneous data are given in Appendix \ref{sec:appendix-filtering}.

\subsection{Evaluation Metrics}
\label{sec:evaluation-metrics}

% For the convenience of comparing to existing work, we use similar evaluation metrics as \citet{serac}. 

\paragraph{Edit Success (ES)}

An edit is successful when all examples in its edit scope are correctly predicted, so we define the edit success of an edit as the accuracy of the model on the edit scope.

\paragraph{Behavior Preservation (BP)}

An input with no relevant edits should not have its prediction changed. Therefore, we define the behavior preservation of an edit as the proportion of unrelated examples whose behavior has been preserved. %A higher BP is better.
% \footnote{This is equivalent to $1 - \text{DD}$, where $\text{DD}$ is the drawdown defined by \citet{serac}.}

\paragraph{Edit Quality (EQ)}

A good model editor should ensure both ES and BP,
% we wish to penalize naive methods that only ensure high performance on one metric (e.g., the unedited model would be perfect in terms of behavior preservation).
hence, we define the edit quality as the harmonic mean of ES and BP.

A perfect model editor has an ES, BP, and EQ of value 1.

\subsection{Implementational Details}

We apply EREN to edit the publicly available FLAN-T5 \citep{flan2}, which is obtained by multitask instruction-finetuning T5 checkpoints \citep{t5}. 

We use Contriever \citep{contriever} for the rough estimation. It is a dense passage retriever with state-of-the-art zero-shot performance. Unless specified, $k=5$ edits are retrieved during inference. To aggregate the retrieved notes $\mathcal E_R$ for feeding to the model as the context, we simply concatenate the notes with a new line as the delimiter. Answers are generated by greedy search. We cap the number of output tokens at 20 and 10 for QA and FC, respectively.

\subsubsection{Task Reformatting}

During the reading step, QA and fact-checking inputs are reformatted as reading comprehension and NLI, respectively. In reading comprehension, we prompt the reader to output ``unanswerable'' if the context cannot be used to answer the question. In NLI, the retrieved notes are the premise and the input is the hypothesis, and the reader is given three options at the end of the prompt, corresponding to entailment, neutral, and contradiction.\footnote{Unfortunately, the instructions for NLI tasks in the FLAN dataset have no clear distinction between contradiction and neutral. I.e., ``No'' and ``It's impossible to say'' both could imply no entailment.
The author says it was an arbitrary choice: \url{https://github.com/google-research/FLAN/issues/32}. 
Despite so, our method achieves superior results.}

\begin{table*}[t!]
    \small
    \centering
    \begin{tabular}{l|ccc|ccc|ccc}
    % \hline
    \toprule
    % \multicolumn{10}{c}{Manual filtering}\\
    % \hline
     & \multicolumn{3}{c|}{\textbf{QA}} 
     & \multicolumn{3}{c|}{\textbf{FC}} 
     &  \multicolumn{3}{c}{\textbf{FC (clean)}}
    \\
    \textbf{Methods}
    & \textbf{ES $\uparrow$} 
    & \textbf{BP $\uparrow$} 
    & \textbf{EQ $\uparrow$} 
    & \textbf{ES $\uparrow$} 
    & \textbf{BP $\uparrow$} 
    & \textbf{EQ $\uparrow$} 
    & \textbf{ES $\uparrow$} 
    & \textbf{BP $\uparrow$} 
    & \textbf{EQ $\uparrow$} 
    \\
    % \hline
    \midrule
    \textit{Unedited} 
        & 0.0 & 100 & 0.0 
        & 41.9 & 100  & 59.1 
        & 38.7 & 100  & 55.8\\
    % \hline
    \midrule
    \multicolumn{10}{c}{\textit{Non-Black-Box Methods}} \\
    % \hline
    \midrule
    Full FT
        & 17.0 & 0.4 & 0.8 
        & 58.9 & 49.8 & 54.0
        & 58.8 & 13.0 & 21.2\\
    MLP FT
        & 1.9  & 9.2  & 3.1  
        & 58.9 & 49.7 & 53.9 
        & 57.7 & 60.8 & 59.2 \\
    MEND (\citeauthor{mend})
        & 0.0 & 0.0 & 0.0
        & 0.0 & 0.0 & 0.0
        & 0.0 & 0.0 & 0.0 \\
    ROME (\citeauthor{rome})$^*$
        & 17.2 & 7.3 & 8.5
        & - & - & - 
        & - & - & - \\
    % \hline
    \midrule
    \multicolumn{10}{c}{\textit{Black-Box Methods}} \\
    % \hline
    \midrule
    SERAC (\citeauthor{serac})
        & 96.9  & 51.4  & 67.2 
        & 60.1 & 67.9 & 63.8
        & 60.4 & 97.6 & 74.6 \\
    \space + Data Aug. 
        & 93.9  & 75.3 & 83.6 
        & 59.2 & 66.7 & 62.7
        & 58.5 & \textbf{98.1} & 73.3 \\
    One-step MRC    
        & \textbf{98.4} & 24.5 & 39.2 
        & \textbf{90.9} & 64.6 & 75.5
        & \textbf{93.8} & 74.5 & 83.0 \\
    EREN (Ours) 
        & 96.9 & \textbf{96.8} & \textbf{96.9} 
        & 81.5 & \textbf{79.6} & \textbf{80.5}
        & \textbf{93.8} & 96.7 & \textbf{95.2}\\
    % \hline
    \bottomrule
    \end{tabular}

    \caption{\label{citation-guide}
    Comparison of different methods on question answering (QA) and fact-checking (FC). FC (clean) is the manually filtered dataset. The base model in One-step MRC and EREN are FLAN-T5-XL. The best result in each metric is \textbf{bolded}. *: Applied to GPT-2-XL instead because it is only applicable to causal language models.
    }
    \label{tab:main-result}
\end{table*}

\subsection{Experimental Details}

\subsubsection{Black-Box Baselines}

\paragraph{SERAC} We primarily test our model against SERAC \citep{serac}, the state-of-the-art model editor. To the best of our knowledge, it is the only editor that can be used in a black-box setting. The original SERAC is finetuned on supervised data, but we are interested in the zero-shot performance, so we train SERAC on ZsRE using the same hyperparameters as \citet{serac} and evaluate it on unseen datasets without additional finetuning. We also evaluate the version of SERAC that employs data augmentation (DA) by automatically sampling similar inputs using a Sentence-BERT \citep{sentence-bert} as negative samples.

\paragraph{One-Step MRC} Let the model directly answer the question in one forward pass in a zero-shot manner.

\subsubsection{Non-Black-Blox Baselines}

For reference, we also list the results of non-black-box model editors, although they should not be regarded as baselines because they have access to the parameters.

\paragraph{Full FT \& MLP FT} Finetune all parameters or finetune only the second linear layer in one of the FFN layers (choosing the best performing one among all layers). They are trained with a constant learning rate of $1e-5$ with Adam optimizer \cite{adam} until the target output is learned or after 50 parameters update steps.

\paragraph{MEND} MEND~\citep{mend} learn to map the gradients to low-rank parameter updates that better ensure the generality and locality of knowledge editing.

\paragraph{ROME} ROME~\citep{rome} modifies a key-value association in the MLP layers by updating the parameters to maximize the probability of the target text. Since ROME requires knowledge about the subject of an edit, we do not evaluate it on FEVER, which does not have labeled subject entities. The base model for ROME is GPT-2-XL~\citep{gpt2} instead because it is unclear how it can be applied to encoder-decoder models.\footnote{MEMIT~\citep{memit}, an extension of ROME that addresses multiple edits, is not considered because it requires edits to be applied simultaneously, but this paper addresses the sequential model editing problem.}

\subsection{Edit Format}

Some methods (e.g., SERAC and gradient-based methods) require input-output pairs as edits. In such cases, we pick one QA pair from the edit scope. For EREN, edits are assumed to come in the format of declarative statements. Therefore, for each example in QA, we convert one of the QA pairs into a declarative sentence using a T5-3B finetuned on QA-NLI \citep{nli-verify,qa-nli}.

\paragraph{Edit Scheme}

We apply 1024 edits sequentially for auto-filtered QA and FC, and 128 edits for the cleaner FC because it only has 128 examples.

%% file: contents/result.tex
\section{Result}
\label{sec:result}

The result for our method and the baselines on QA and fact-checking are shown in Table~\ref{tab:main-result}. The edit quality of EREN is greater than the non-black-box baselines and SERAC by a large margin. The finetuning methods, MEND, and ROME suffer from severe catastrophic forgetting, resulting in very low edit quality, and generally fail to sequentially apply more than a thousand edits. After a certain number of sequential parameter updates, the model has degraded to producing unintelligible text. For MEND, it is because the hypernetwork was adapted to the parameter of the base model, but the parameter changes for each update, while the hypernetwork stays the same. A similar problem is found in ROME, where we have to pre-compute the activation statistics of the base model, which is not updated for each edit.\footnote{It is prohibitively expensive to update the hypernetwork and activation statistics after every edit.} It is also worth noting that the time needed to apply each edit in these non-black-box baselines is significantly more than SERAC and EREN.

Interestingly, one-step MRC can beat SERAC in fact-checking in terms of editor quality. This is likely because SERAC is trained on a QA dataset and is therefore unable to adapt to the domain of fact-checking in a zero-shot manner. 

One of the main reasons SERAC underperforms EREN is that SERAC is limited by the two small complementary models.
See Appendix~\ref{appendix:comparison-to-serac} for a discussion on why SERAC underperforms.
% The appendix will include a discussion on why SERAC underperforms.

% Perhaps surprisingly, data augmentation with similar questions as negative samples does not result in better performance in fact-checking for SERAC.

% Additionally, although non-black-box methods have access to the parameters of the base model, they generally fail to make more up to thousands of edits (more details in Section~\ref{sec:different-number-of-edits}).

In the following sections, we will evaluate EREN's performance in editing different base models, scaling the number of edits, and its ability to combine multiple edits. We also analyze the effect of the note retriever. Finally, we show that the hard in-scope and out-of-scope examples are more challenging than those that are commonly used to test model editors.

\subsection{Different Base Models}

\begin{figure}
    \centering
    \includegraphics[width=\linewidth]{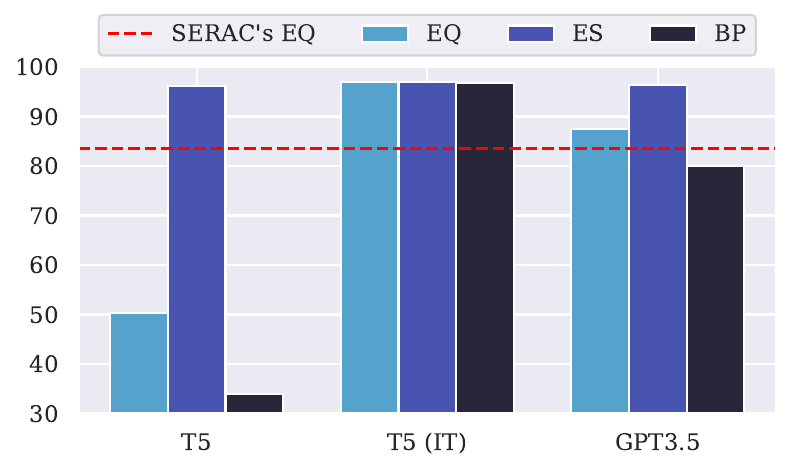}
    \caption{The performance of EREN on different base models. The red dotted line represents the EQ of SERAC + DA. T5 and T5 (IT) are the non-instruction-tuned and instruction-tuned versions of T5-XL, and GPT3.5 is the \texttt{gpt-3.5-turbo} API. 
    See Appendix~\ref{appendix:different-base-models} for more details.
    % More details will be given in the appendix.
    }
    \label{fig:diff_base_models}
\end{figure}

Figure~\ref{fig:diff_base_models} shows the result of EREN on different base models, the implementation details are given in Appendix \ref{appendix:different-base-models}. 
We can see that instruction-tuning is crucial for the success of EREN, i.e., T5 without instruction-tuning~\citep{t5} has only half the edit quality. Interestingly, most performance degradation comes from low BP, which indicates that instruction-tuning is essential for ensuring the LLM is robust to irrelevant contexts. 

We also observe that EREN is effective for editing GPT3.5, an API-level LLM. It is not as effective as using T5 because GPT3.5 is trained to output chat-like responses, which gives a lower score on QA because we use exact match as the evaluation metric.

\subsection{Different Number of Edits}

\label{sec:different-number-of-edits}

\begin{figure}[t!]
    \centering
    \includegraphics[width=\linewidth]{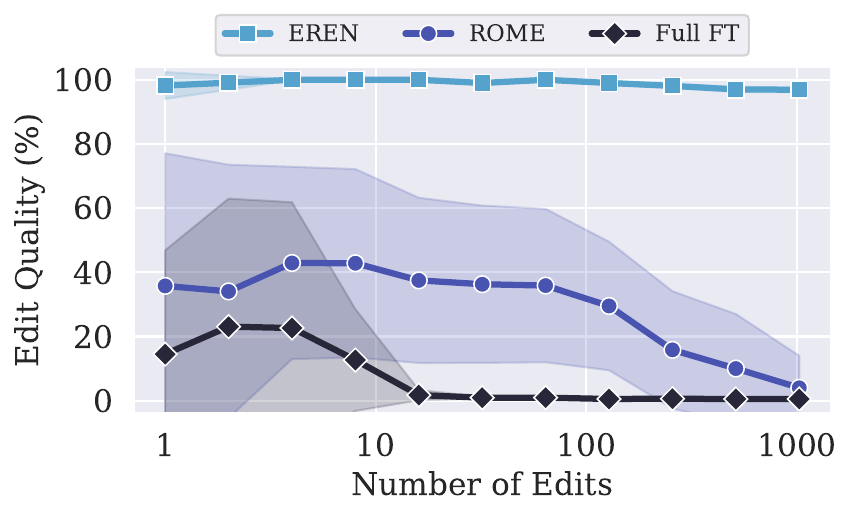}
    \caption{Edit quality of EREN, ROME, and Full FT by different numbers of edits on \textsc{CounterFact}. The colored area is the standard deviation of 5 runs.}
    \label{fig:number-of-edits}
\end{figure}

% It is desirable to be able to apply a large number of edits without decreased performance. 
\citet{memit,serac} have shown that existing methods may struggle when we scale up the number of edits. 
Figure~\ref{fig:number-of-edits} shows the performance of EREN, ROME, and Full FT on question-answering where the number of edits ranges from 1 up to 1024. 
Both Full FT and ROME exhibit very poor EQ at 1024 edits because sequentially editing the model's parameters will add up the errors of each edit.
In contrast, EREN keeps the base model frozen, making the impact of each edit limited to relevance estimation, and reducing the EQ drop to less than 4\%.

\subsection{Combining Multiple Edits}

\begin{table}[t!]
    \small
    \centering
    \begin{tabular}{l|ccc}
    % \hline
    \toprule
    \textbf{Method} 
    & \textbf{ES $\uparrow$} 
    & \textbf{BP $\uparrow$}
    & \textbf{EQ $\uparrow$} \\
    % \hline
    \midrule
    \textit{Unedited} & 23.4 & 100 & 37.9 \\
    EREN ($k=1$)  & 38.9   & \textbf{95.7} & 55.3 \\
    EREN ($k=2$)  & 60.9   & 93.2 & 73.7 \\
    EREN ($k=3$)  & 65.4   & 92.0 & 76.5 \\
    EREN ($k=5$)  & 67.2   & 89.5 & 76.7 \\
    EREN ($k=10$) & \textbf{70.7} & 89.1 & \textbf{78.8} \\
    % \hline
    \bottomrule
    \end{tabular}
    \caption{
        Performance on questions that require combining knowledge from multiple edits. $k$ is the number of retrieved notes.
    }
    \label{tab:combine-multiple-edits}
\end{table}

% In practice, an input may be relevant to multiple edits. 
Existing retrieval-based methods~\citep{serac,grace} assume that each input has only one relevant edit, and they are not able to combine multiple edits. To evaluate the ability of our method to combine knowledge from multiple edits, we sample 512 examples from HotpotQA~\citep{hotpotqa} and insert all passages as edits, then sample another 512 examples to use as out-of-scope questions. The result is listed in Table~\ref{tab:combine-multiple-edits}. We can see that the performance increases sharply when retrieving more than one edit, which indicates the ability to combine multiple edits.
% However, the edit quality is still significantly lower than QA and fact-checking. 
% It is because the passages consist of multiple sentences, which may create noise for both the retriever and the reader. However, in practice, it is reasonable to expect that edits are provided by the maintainers of the LLM, and are therefore formatted as concise single sentences. 

% We also observe a significant decrease in behavior preservation with more retrievals. Intuitively, more retrievals make it more likely for the reader to incorrectly classify some notes as relevant. 

% \subsection{Edit Quality by Model Size}

% \begin{figure}[t!]
%     \centering
%     \includegraphics[width=\linewidth]{images/eq_by_reader_size.pdf}
%     \caption{Edit quality on QA by different reader model sizes. SERAC + DA (the second best method) is shown plotted with dotted lines.}
%     \label{fig:reader}
% \end{figure}

% The EQ by different model sizes is shown in Figure \ref{fig:reader}. EREN's performance increases steadily with size, topping at around 3B parameters because it has reached near perfect. It is reasonable to expect that the EQ will scale beyond 3B parameters for scenarios with more difficult relevance estimation (e.g., complex reasoning), where the current reader size is not sufficient. 
% SERAC with data augmentation is only able to beat EREN when the reader is less than 220M parameters.

\subsection{Effect of Note Retriever}

% \begin{table}
%     \centering
%     \begin{tabular}{l|cccc}
%     \hline
%     $k$ 
%     & \textbf{Recall} 
%     & \textbf{ES}
%     & \textbf{BP}
%     & \textbf{EQ} \\
%     \hline
%     1  & 98.90 & 90.1 & 98.6 & 94.1 \\
%     2  & 99.72 & 92.9 & 97.9 & 95.4 \\ 
%     3  & 99.75 & 95.8 & 97.4 & 96.6 \\
%     5  & 99.75 & 96.9 & 96.8 & 96.8 \\
%     10 & 99.82 & 97.1 & 96.3 & 96.7 \\
%     25 & 99.90 & 97.3 & 95.9 & 96.6 \\ 
%     \hline
%     \end{tabular}
%     \caption{
%     Edit quality with different number of retrievals $k$ on QA. Precision is not so important as the reader will be able to discern unrelated edits.}
%     \label{tab:retrieval}
% \end{table}

\begin{figure}[t!]
    \centering
    \includegraphics[width=\linewidth]{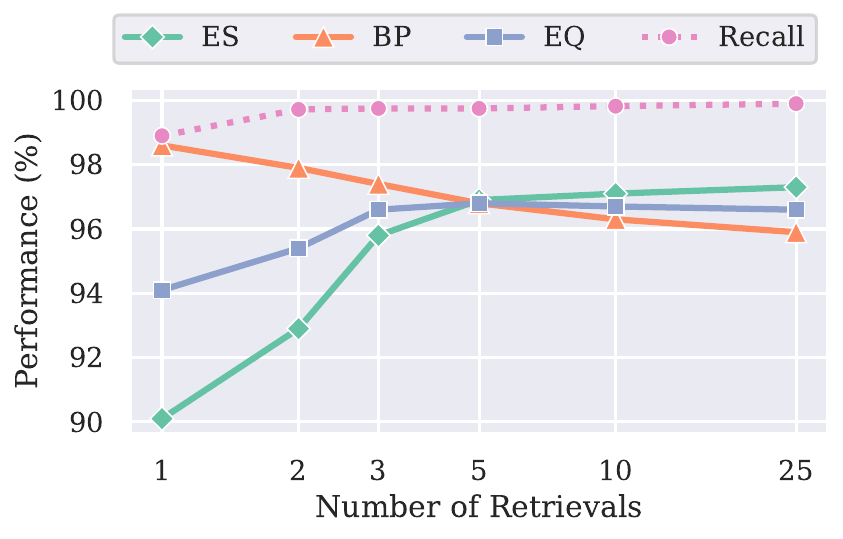}
    \caption{The performance of EREN and recall rate of the note retriever by retrieving different numbers of notes on \textsc{CounterFact}.}
    \label{fig:num_retrieval}
\end{figure}

A note retriever filters out highly dissimilar notes to speed up inference, and it may significantly influence the final performance. 
Figure \ref{fig:num_retrieval} plots the retrieval recall rate and EREN's performance on \textsc{CounterFact} with varying numbers of retrievals. 
% From it, we see that the precision of the note retriever has a very limited impact on the final performance. 

Interestingly, with a small number of retrievals, the ES is significantly lower than the recall rate, which means that although the reader can see the relevant note, it has not been able to produce the correct answer. 
We hypothesize that this is because the reader is instruction-tuned on datasets where contexts are usually longer than just a few sentences, and struggles to generalize to shorter contexts. 
% Consequently, the model struggles to generalize to our situations where the context consists of only a very few numbers of sentences. 

On the other hand, increasing the number of retrievals reduces BP, which is intuitive, because more irrelevant notes may introduce noise for the reader. The edit quality does not increase much beyond 5 retrievals. 

% It is important to note that the recall rate may be lower if the number of edits is sufficiently large. However, in such cases, simpler finetuning methods (with appropriate constraints) can be applied \citep{memit}.

\subsection{Harder Out-of-Scope Examples}

\begin{table}
    \small
    \centering
    \begin{tabular}{l|cc}
    % \hline
    \toprule
    \textbf{Method} 
    & \textbf{NB. Subj.} $\uparrow$
    & \textbf{Same Subj.} $\uparrow$ \\
    % \hline
    \midrule
    % Full FT      & 0.0    & 0.0 \\
    SERAC        & 89.3 & 33.1 \\
     + Data Aug. & 89.4 & 63.4 \\
    One-step MRC & 13.7 & 25.1 \\
    EREN       & \textbf{97.0} & \textbf{95.9} \\
    % \hline
    \bottomrule
    \end{tabular}
    \caption{
    Behavior preservation on \textsc{CounterFact} by different types of out-of-scope questions. \textbf{NB. subj.}: Questions where the subject is replaced with a ``neighbor subject'', introduced by \citet{rome}. \textbf{Same subj.}: Questions about unrelated knowledge of the same subject as the edit, introduced by us.}
    \label{tab:result-hard-oos}
\end{table}

As mentioned in Section~\ref{sec:dataset-qa}, although \textsc{CounterFact} already includes hard out-of-scope questions on neighboring subjects, we collect unrelated questions about the subject of the edit, which serves as harder out-of-scope questions. Table~\ref{tab:result-hard-oos} shows the breakdown of behavior preservation of different methods on the two types of out-of-scope questions. We observe that SERAC sees a much larger performance drop on out-of-scope questions about the same subject compared to the out-of-scope questions in the original \textsc{CounterFact}, which confirms our hypothesis that the question we collect about the same subject is more challenging for model editors than the questions in the original dataset. This is likely because SERAC's scope classifier has learned to overly rely on the subject as a signal to determine the relevance of edits. Using negative samples as data augmentation significantly mitigates this, but still is far behind the performance of EREN.

\subsection{Harder In-Scope Examples}

\citet{serac} proposed to construct hard in-scope QA by automatically constructing implied facts. E.g., the QA pair (``Who is the Prime Minister of UK?'', ``Boris Johnson'') as an edit would imply the QA pair (``Where is Boris Johnson Prime Minister?'', ``UK''). However, we discover that SERAC fails on simple in-scope examples unseen in its training set, such as rephrasing the question as a boolean question. Concretely, after applying the above edit and asking ``Is it true that Boris Johnson is the Prime Minister of the UK?'', SERAC would still output ``Boris Johnson''. We randomly sample 512 edits from \textsc{CounterFact} and convert them into boolean questions with the prompt ``Is it true that \{edit statement\}?''. 

The result is shown in Table~\ref{tab:boolq}. SERAC gets all questions wrong. We conclude this is because the counterfactual model that is responsible for inference on in-scope examples is too small\footnote{SERAC finetunes a T5-small for the counterfactual model.}, and has overfitted to the training data that includes automatically generated implied examples.

\begin{table}
    \small
    \centering
    \begin{tabular}{l|cccc}
    \toprule
    \textbf{Method} & Unedited & Full FT & SERAC & EREN \\ \midrule
    \textbf{ES $\uparrow$} & 9.4 & 4.7 & 0.0 & 100 \\
    \bottomrule
    \end{tabular}
    \caption{
    Edit success on simple boolean questions converted from \textsc{CounterFact} with the template ``Is it true that \{edit statement\}'', which act as harder in-scope examples.}
    \label{tab:boolq}
\end{table}

%% file: contents/conclusion.tex
\section{Conclusion}    

This work has presented EREN, a zero-shot retrieval-based model editing framework for black-box LLMs that is scalable. The editor stores all edits in a growing notebook in natural text, and a reader uses the notes to produce an answer if any of them are relevant. Our experiments were conducted under a black-box setting, where access to datasets of edits and model parameters and activations was not available, and we tested the model's ability to combine multiple edits, increasing its practicality and applicability to a broad range of scenarios. Our empirical results show that EREN significantly outperforms the current state-of-the-art model editor, demonstrating superior edit success and preservation of the model's behavior on unrelated edits. We believe that EREN represents a significant step towards the lifelong maintenance of LLMs.

% \section*{Limitations}

% It is important to note that EREN does have limitations, specifically its impact on inference time. The two inference steps, coupled with a potentially lengthy context reading, can result in a significant increase in processing time. Additionally, as edits are continually applied, the notebook (and index for note retrieval) may grow indefinitely and the reader's performance may deteriorate.  To address these concerns, future research efforts could explore more effective methods for composing shorter contexts and transitioning to alternative editors as the number of edits increases. The performance of EREN is also sensitive to the exact wording of the prompt, and designing prompts in a more scientific manner may mitigate this.

% Additionally, many aspects of the model editing problem are still unaddressed by most of the existing works, including this paper. For instance, how the updated knowledge transfers across languages, how effective the editors are on specialized domains, etc.

\section*{Ethics Statement}

The ability to quickly update knowledge in LLMs has many benefits, but may also be used to inject wrong knowledge, undesired behavior, or bias into LLMs, although that is not the motivation of our work. This method could also significantly increase the input length, which may result in a higher carbon footprint. However, we emphasize that the purpose of editing models is to avoid the more expensive choice of re-training the model when an update to the parametric knowledge is requested. It is also worth noting that most existing model editing methods introduce more computation and memory overhead than our method.

%% file: contents/appendix.tex
\appendix

\section{Dataset Construction Details}
\label{appendix:dataset-construction}

\subsection{Question Answering: \textsc{CounterFact}}

Our version of \textsc{CounterFact} includes harder out-of-scope questions in which the relation and object of the edit's triple are changed, but the subject stays the same, the resulting questions are questions that elicit knowledge about the subject of the edit, but are outside the edit scope. To construct such questions, we query triples from Wikidata using the subject name. We use the templates in \textsc{CounterFact} to verbalize the triples (discarding those without a template) into declarative sentences. Then we convert the templates to QA pairs using MixQG\footnote{\url{https://huggingface.co/Salesforce/mixqg-large}}. MixQG accepts a context and an answer as the input and produces a question, and since each \textsc{CounterFact} template is supposed to prompt the answer as a text continuation, we simply append the answer to the template, and feed this as the context to MixQG.

To create an edit, we convert one QA pair from the edit scope into a declarative sentence with a pretrained converter\footnote{\url{https://huggingface.co/domenicrosati/QA2D-t5-base}}, which is a T5-Base finetuned on QA2D \citep{qa-nli}. Interestingly, although we could have simply appended the answer to the prefix prompt provided in the vanilla \textsc{CounterFact}, we found that converting it to a QA pair and then to a declarative sentence may eliminate ambiguity in the \textsc{CounterFact} prompts.

Note that the edit scope in \textsc{CounterFact} is created by simply paraphrasing one question, which means all answers are the same. 
An ideal edit scope for evaluation should include all QA pairs implied by the edit, but such data is too expensive to collect. 
Thus, we leave such study for future work.

\subsection{Fact-Checking: FEVER}

As mentioned in Section \ref{sec:fever}, we use the version of FEVER that is released by \citet{decao-editing}, which includes paraphrases created with back-translation. However, when we want to edit the model to make falsify a fact, we need to construct the negation of the factual statement. To do so, we turn the statement into a boolean question with ``Is it true that {statement}?'' and feed this question along with ``no'' to the QA pair to the statement converter as mentioned above.

\subsection{Filtering}
\label{sec:appendix-filtering}

As mentioned in Section \ref{sec:filtering}, we create cleaner versions of \textsc{CounterFact} and FEVER by removing poorly worded or labeled examples. Table \ref{tab:bad-example-example} lists some examples for each type of bad example. To reduce the computational cost, we first use a pretrained NLI model \citep{anli} to automatically construct a larger filtered dataset, then create a smaller but cleaner version by manual filtering.

\subsubsection{Automatic Filtering}

We set the edit statement as the premise, then feed each in-scope and out-of-scope input as the hypothesis to the NLI model. But since \textsc{CounterFact} examples are QA-pairs, we first convert them to statements using a converter\footnote{\url{https://huggingface.co/domenicrosati/question_converter-3b}} trained on QA2D \citep{qa-nli}.

\begin{itemize}
    \item For \textbf{in-scope inputs}, we regard it as incorrectly labeled if the edit statement is neutral to the input (neither entails nor contradicts). However, since the NLI model is imperfect, we require that the predicted probability of neutral be less than 80\% of the self-entailment probability of the edit statement (i.e., the probability of the statement entailing itself).
    \item For \textbf{out-of-scope inputs}, we want the edit statement to be neutral to the inputs. Again, we require that the probability of the edit statement is neutral to the input to be greater than 80\% of the self-entailment probability of the edit statement.
\end{itemize}

It is important to note that inputs with unintelligible wording may not have been taken care of with this filtering process. This is because unintelligible wording of a fact is likely to be neutral of other facts, so out-of-scope inputs with unintelligible wording are likely to survive this filtering process.

Moreover, out-of-scope input of an edit may be relevant to another edit, and such false out-of-scopes are not taken care of with this procedure. In \textsc{CounterFact}, such examples are rare since different edits have different subjects, but in FEVER, for example, there is an edit statement ``Saxony is in Ireland'' and an out-of-scope input ``Saxony is the sixth most populous Spanish state''. Performing NLI on all pairs of edit statements and out-of-scope inputs would be too expensive, therefore, we keep them for the larger filtered data and filter them out by manual filtering.

For \textsc{CounterFact}, the most common errors are false out-of-scope and false in-scope questions. False out-of-scope questions mainly arise from the verbalized triples from Wikidata. This is because many templates in \textsc{CounterFact} are about closely related facts, and the templates are too ambiguous to distinguish the differences. For instance, a template about a person's city of birth may be mistaken for being about the person's country of birth, such as ``Where was \{subject\} born?''.

\begin{table*}[htb!]
    \centering
    \begin{tabular}{p{32mm}|p{116mm}}
        \toprule
        \textbf{Type} & \textbf{Example} \\
        \midrule
        \multicolumn{2}{c}{\textsc{CounterFact}} \\
        \midrule
        Bad wording
            & Edit: ``Toko Yasuda, the'' \\
        \midrule
        False in-scope 
            & Edit: ``Danielle Darrieux's mother tongue is English.'' \\
            & In-scope question: ``What is the nationality of Danielle Darrieux?'' \\ 
        \midrule
        False out-of-scope 
            & Edit: ``Danielle Darrieux's mother tongue is English.''\\
            & Out-of-scope question: ``What language does Danielle Darrieux speak?'' \\
        \midrule
        \multicolumn{2}{c}{FEVER} \\
        \midrule
        Bad wording
            & In-scope fact: ``==References====External links=='' \\
        \midrule
        False in-scope 
            & Relevant Edit: ``Nicholas Brody is a character on Homeland.'' \\
            & In-scope fact: ``Nicholas Brody is a character at home.'' \\ 
        \midrule
        False out-of-scope 
            & Relevant Edit: ``Jayasudha is an actor that stars in Daag''\\
            & Out-of-scope fact: ``Daag is a painting'' \\
        \bottomrule
    \end{tabular}
    \caption{Examples of the different types of error found in \textsc{CounterFact} and FEVER. Note that the examples of \textsc{CounterFact} here is after mining hard out-of-scope examples and prefix-to-question conversion, but most of these errors are found before conversion as well.}
    \label{tab:bad-example-example}
\end{table*}

\subsubsection{Manual Filtering}
\label{sec:manual-filtering}

The automatically filtered version of FEVER still contains some erroneous examples as shown in Table \ref{tab:bad-example-breakdown}. Therefore, we pick the first 128 examples from FEVER and manually filter out those that are unintelligibly worded or are too ambiguous. Since we use another 128 examples as out-of-scope examples, they must be unrelated to the former 128 edits. Therefore, to filter out false out-of-scopes, we have to read the 128 edits, and make sure every out-of-scope examples are irrelevant to each of the 128 edits.

\section{Prompts}
\label{appendix:prompts}

When the reader in EmoRen determines that there are no relevant edits, it outputs a predefined string on irrelevant edit context. This predefined string is usually specified in the prompt. The prompts that we use for QA and fact-checking are listed in Table \ref{tab:prompts}, where the string for irrelevance is ``unanswerable'' and ``It's impossible to say'' respectively. However, we find that the wording of this string has little impact on the performance of our preliminary experiments as long as they are semantically equivalent.

% 表格：不同任务的 Prompts

\begin{table*}
    \centering
    \begin{tabular}{p{24mm}|p{126mm}}
        \toprule
        \textbf{Task Type} & \textbf{Prompt} \\
        \midrule
        MRC 
        & Read this and answer the question. If the question is unanswerable, 
say "unanswerable".
\\ \\
& <context>
\\ \\
& <question>
\\

\midrule
QA without context
        & Please answer this question: <question>\\
        \midrule
        Fact-checking with context 
        & <context>
\\ \\
& Based on the paragraph, can we conclude that "{hypothesis}"?
\\ \\
& OPTIONS:

- Yes

- It's impossible to say

- No
\\
        \midrule
        Fact-checking without context 
        & Is it true that <hypothesis>?
        \\
        \bottomrule
    \end{tabular}
    \caption{The prompts that we used for QA and fact-checking, which are hand-picked from the instructions in the FLAN collection with slight modification.}
    \label{tab:prompts}
\end{table*}

\section{Additional Experimental Details}

\subsection{ROME}

In this work, we applied ROME on GPT-2-XL instead of T5, because they can only be applied to causal LMs. For better reproducibility, we used the publicly released pre-computed layer statistics to edit the pretrained GPT-2-XL. We use the following prompt with few-shot exemplars to guide the model to perform QA.

\begin{small}
    
\begin{verbatim}
Q: Who is the President of China?
A: Xi Jinping

Q: When did World War II end?
A: 1945

Q: What is the capital of Norway?
A: Oslo

Q: Who is the main character in The Matrix?
A: John Wick

Q: Who is the founder of Apple?
A: Steve Jobs

Q: Which is the largest planet in our solar system?
A: Jupiter

Q: How many legs do spiders have?
A: Eight

Q: Does pure water conduct electricity?
A: No

Q: {question}
A:
\end{verbatim}
\end{small}

Despite using this prompt, the model still displays a strong tendency to output additional text after the answer. Therefore, we only regard the first line of the output sequence as the answer.

% \subsection{In-Context Learning}

% Our in-context learning baseline is similar to IKE~\citep{ike}, but we have fewer number of demonstrations. IKE has 32 demonstrations, but they assume that only one edit has been applied to the model, which means there is only one example whose relevance needs to be estimated. In contrast, our experimental setting scales up to 1024 examples, and we retrieve 5 edits during inference, there edits need to be included in every demonstration, resulting in a much lengthier context. Therefore, we reduce the number of demonstrations to 8, which represent a much more realistic use case of in-context learning for knowledge editing. 

\section{Different Base Models}
\label{appendix:different-base-models}

We use few-shot demonstrations for T5 and GPT3.5, because we find that these models (the former is non-instruction-finetuned and the latter is fine-tuned for chatting) cannot reliably follow the instructions. Specifically, they are inclined to produce much more tokens in addition to the actual answer, such as explanation for the context is relevant/irrelevant.

\section{Comparison to Evaluation Metrics in Existing Works}

The original paper of ROME~\citep{rome} only evaluated the ability to apply one edit. \citet{memit} showed, through empirical results, that the \textit{efficacy} of ROME drops steadily with increased number of edits. Here, the ``efficacy'' metric is defined as the proportion of in-scope examples where the probability of the target answer is greater than the probability of the actual answer, i.e., $ \mathbb{E} [P(y^{*}) > P(y)]$, where $y$ and $y^*$ are the original and post-edit target outputs. However, our evaluation metric, edit success, is more challenging, because only the exactly matches of the top-1 output sequence are counted as successful edits. 

\section{Comparison to SERAC}
\label{appendix:comparison-to-serac}

EREN differs from SERAC in the following points.

\begin{itemize}
    \item \textbf{Eliminating the need to train an extra in-scope model.} SERAC needs to train a counterfactual model to use as the in-scope model. Instead, EREN demonstrates how to leverage the reading comprehension capabilities of LLMs to perform editing as the in-scope model and directly use LLMs as the out-of-scope model. In other words, we are unifying the in-scope and out-of-scope models into one single model. One of the main reasons SERAC underperforms EREN is that the small in-scope model has a different parametric memory (because it is much smaller) than the base model, therefore, when the relevance estimation fails and an edit is falsely identified as being relevant, the in-scope model will not be able to ignore the edit and produce the same prediction as the base model.
    \item \textbf{Improving relevance estimation.} SERAC has two kinds of relevance estimation. The first one is to use a dual-encoder to calculate the distance between the encodings of the question and edits as their relevance, which is less capable (this is the primary method used in SERAC's paper and our paper). The second one is to iterate through every edit, concatenate the edit with the question, and feed them into a binary classifier, which is very slow.  In EREN, we design a two-step retrieval process. EREN first has a rough estimation process that eliminates highly dissimilar edits, and then the reader can condition on multiple edits simultaneously to estimate the relevance of edits. This method is both expressive and efficient.
    \item \textbf{Supporting edit combination.} If one input is relevant to multiple edits, SERAC is not able to combine knowledge from the edits to produce a correct answer. This is because the counterfactual model is trained to condition on one input-output pair at a time. In contrast, EREN achieves this by performing generation conditioned on all edits (the entire notebook). Passing only top-1 most relevant edit to the counterfactual model also means that SERAC would very likely produce wrong answers whenever the top-1 prediction of the scope classifier is incorrect.
\end{itemize}

% \subsection{Editing Dynamics}

% When using the more challenging edit success, we discover an interesting sudden drop of edit success of ROME around 175 edits.

% \end{appendices}